\documentclass{article}

\usepackage{arxiv}
\newcommand{\undertitle}{}
\newcommand{\headeright}{}
\usepackage{natbib}
\usepackage{amsmath}
\usepackage{amssymb}

\usepackage[utf8]{inputenc} 
\usepackage[T1]{fontenc}    
\usepackage{hyperref}       
\usepackage{url}            
\usepackage{booktabs}       
\usepackage{amsfonts}       
\usepackage{nicefrac}       
\usepackage{microtype}      
\usepackage{lipsum}
\usepackage{graphicx}
\graphicspath{{./images/} }

\title{Retrieval-Augmented Generation vs. Deterministic Tax Computation in Multi-Agent Financial Advisory: A 2×2 Factorial Experiment }

\author{
 Aryan Brar \\
  {\footnotesize Royal Bank of Canada}\\
  {\footnotesize\texttt{abrar82@uwo.ca}} \\
   \And
 Justin Du\\
  {\footnotesize Royal Bank of Canada}\\
  {\footnotesize\texttt{justin.du@mail.utoronto.ca}} \\
  \And
 Avery Lor \\
  {\footnotesize Royal Bank of Canada}\\
  {\footnotesize\texttt{avery.lor@mail.utoronto.ca}} \\
  \And
 Kylie Seto\\
  {\footnotesize Royal Bank of Canada}\\
  {\footnotesize\texttt{ksa210@sfu.ca}} \\
  \And
 Eric Taylor\\
  {\footnotesize RBC Borealis}\\
  {\footnotesize\texttt{eric.j.taylor@borealisai.com}} \\
}
\date{}

\begin{document}
\maketitle
\vspace{-0.35in}
\begingroup
\renewcommand{\thefootnote}{}
\footnotetext{This work was completed at Royal Bank of Canada as part of the RBC Amplify program.}
\endgroup
\begin{abstract}
Tax-loss harvesting demonstrates consistent benefits to long-term portfolio growth; yet implementing it efficiently often involves complex considerations that are specific to the holdings within that portfolio and the individual who owns it. We introduce a custom capital gains calculation engine and a RAG-retrieved vector store of market advisory reports to provide context for a multi-agent trade recommendation system. We investigate the effects of each context provider on the quality of recommendations, measured by relative capital gains incurred during portfolio liquidation. A 2\texttimes 2 repeated-measures ANOVA revealed a significant main effect of the tax optimization engine ($F(1,29) = 9.17$, $p = .005$, $\eta^2_p = .240$): enabling the engine \emph{reduced} tax savings by approximately 55 percentage points relative to the no-engine conditions. The RAG main effect was not significant ($p = .841$), nor was the interaction ($p = .553$). The RAG-only condition achieved the highest descriptive mean tax savings (47.7\%), and the baseline condition performed second-best (30.6\%), suggesting that the pre-trained language model's internalized financial knowledge may be sufficient for competent tax-loss harvesting recommendations without explicit tooling. These results indicate that augmenting LLM agents with domain-specific computation engines does not guarantee improved performance and may introduce conflicting optimization signals.
\end{abstract}

\section{Introduction}
\subsection{AI in Wealth Management}
The intersection of artificial intelligence and wealth management has emerged as a transformative domain in financial services. Machine learning and deep learning techniques have progressively enhanced portfolio optimization, risk assessment, and investment decision-making at both institutional and retail scales. Recent advances have moved beyond simplistic rule-based systems toward sophisticated neural architectures capable of learning complex market dynamics from high-dimensional data. Robo-advisors and AI-driven portfolio platforms have democratized access to quantitative investment strategies previously available only to institutional investors with substantial capital and technical infrastructure. However, the deployment of these systems raises critical questions about their effectiveness in real-world conditions, particularly when operating under regulatory constraints and competing objectives such as after-tax performance, transaction cost minimization, and investor alignment.

\subsection{Tax-Loss Harvesting and Tax Alpha}
\citet{berkin2003tax} describe Tax-loss harvesting (TLH) as the practice of selling shares below the original cost to generate tax credits. Tax credits can be used to subtract from a persons’ capital gains in the current year or use to defer gains in future years. TLH represents one of the most direct mechanisms for improving after-tax portfolio returns. Tax alpha as described by Berkin, A. L and Ye, J is the “tax consequences of active management”. It represents the money added to or subtracted from a portfolio due to how capital gains, losses, and other tax events are managed and realized.

\subsection{Multi-Agent Architectures in AI Systems}
Mixture-of-Experts (MoE) and multi-agent architectures have emerged as powerful paradigms for handling heterogeneous, multi-objective problems in machine learning. Rather than forcing a single monolithic model to optimize conflicting objectives, MoE systems decompose the problem space into specialized expert agents, each trained on a distinct objective or task domain, with a learned router or gating mechanism determining expert allocation at inference time. \citet{pishehvar2026three} demonstrated this approach in portfolio management by constructing four specialized experts (momentum, growth, defensive, tax-aware), each optimizing distinct investment mandates, with an intent router blending expert outputs based on active objective and market regime. This architecture naturally accommodates the heterogeneity inherent in personalized portfolio management: different investors have different time horizons, tax brackets, liquidity needs, and behavioral preferences. A reality that monolithic systems struggle to capture. Multi-agent systems also offer interpretability advantages; routing decisions can be examined to understand how objectives are being weighted, and expert specialization can be validated independently. However, the literature on multi-agent portfolio systems remains sparse, and critical questions remain about how to effectively coordinate agents when objectives are partially aligned, conflicting, or unknown at design time. 

\subsection{Retrieval-Augmented Generation and Knowledge Retrieval}
Retrieval-Augmented Generation (RAG) has emerged as a powerful pattern for augmenting language models and decision-making systems with grounded, external knowledge. Rather than relying solely on learned parameters, RAG systems retrieve relevant documents, data, or context from a knowledge base at inference time, using that context to condition generation or decision-making. This approach has demonstrated effectiveness in reducing hallucination in large language models, improving factuality, enabling dynamic knowledge updates without retraining, and providing verifiable justification for decisions. In the financial domain, RAG principles could enhance portfolio management by retrieving relevant market news, regulatory guidance, tax documentation, and domain-specific insights to inform allocation decisions. The potential synergy between RAG and multi-agent systems is significant but underexplored: a tax agent could retrieve relevant tax code sections and recent rulings; a momentum agent could retrieve recent earnings announcements and analyst sentiment; a risk agent could retrieve historical volatility regimes and tail-risk scenarios. However, the integration of retrieval-based knowledge with learned policies, particularly in sequential decision-making contexts like portfolio rebalancing, remains largely unexplored in the academic literature. 

\subsection{Research Gap}
Existing literature has addressed AI in portfolio management, tax-loss harvesting optimization, and multi-agent architectures largely in isolation. Few if any studies have examined the synergistic integration of these three components within a unified system. Specifically: (1) most tax-aware portfolio systems operate deterministically or use conventional optimization without deep learning; (2) multi-agent RL systems in finance typically optimize a single objective (e.g., Sharpe ratio) and do not systematically integrate tax reasoning; (3) retrieval-augmented approaches have not been tested in the portfolio management context to determine whether grounding decisions in external knowledge meaningfully improves performance; and (4) no empirical comparison exists between pure learned policies, knowledge-augmented policies, tax-specialized architectures, and their combinations. The literature provides no principled guidance on whether, and under what conditions, combining these approaches yields additive, synergistic, or diminishing returns. This gap is particularly consequential for retail investors and fintech platforms, where after-tax returns directly translate to user value, and where the practical integration of multiple AI techniques must operate within computational and regulatory constraints. 

\section{Related Works}
\subsection{Tax-Loss Harvesting: Mechanisms and Effectiveness }
Tax-loss harvesting has become an increasingly important strategy for improving after-tax returns in portfolio management. \citet{chaudhuri2020empirical} provided an empirical evaluation of tax-loss-harvesting alpha using historical US equity data from 1926 to 2018, demonstrating that tax-loss harvesting yields an average of 1.08\% annualized alpha before transaction costs, declining to 0.82\% when constrained by wash-sale rules. The study revealed that tax alpha varies substantially across market regimes, performing best during high-volatility periods with low overall returns (such as the Great Depression era, yielding 2.13\% annually) and modestly during low-volatility expansion periods (0.51\% during 1949–1972). 

\citet{israelov2022optimized} advanced the theoretical understanding of tax-loss harvesting by characterizing the optimization problem as a classic risk-reward tradeoff. They established that harvesting efficiency depends on the security's volatility and introduced an efficient frontier between harvesting yield and active risk. Their analysis demonstrated that a -10\% harvesting threshold (monthly) or -15\% threshold (daily) strikes a reasonable balance, and they proposed a novel "throttled harvesting algorithm" for large positions that restricts maximum active weight deviations while maintaining adequate harvesting opportunities. Notably, they found that daily harvesting provides minimal marginal improvement over monthly harvesting when transaction costs are considered. 

\subsection{Tax-Aware Factor Investing and Long-Short Strategies}
\citet{krasner2023loss} examined the mechanisms underlying tax-aware long-short factor strategies, resolving an apparent paradox: how strategies can simultaneously achieve cumulative net capital losses exceeding 100\% of invested capital while maintaining significant pre-tax alpha. Their key finding was that net capital losses in tax-aware long-short strategies arise primarily from gain deferral rather than increased loss realization, a distinction they argue is critical for investor understanding. The study demonstrated that tax-aware strategies achieve this through liquidating loss positions and creating new positions while minimizing gain realization, particularly by deferring short-term gains on long positions. Importantly, they showed that most portfolio turnover remains directed toward the alpha model, explaining the sustained pre-tax performance. 

\subsection{AI-Based and Deep Learning Approaches to Portfolio Management}
\citet{pishehvar2026three} introduced a three-phase deep reinforcement learning system for personalized, tax-aware portfolio management that addresses key limitations of prior work. Phase 1 employs self-supervised learning with a cross-asset encoder augmented by the Chronos time series foundation model, addressing "ticker lock-in" through a 50-dimensional observable metadata vector that generalizes to any publicly traded asset without retraining. Phase 2 introduces a Mixture-of-Experts (MoE) architecture with four specialized expert heads (momentum, growth, defensive, tax-aware) and a learned intent router that simultaneously serves six distinct investment objectives. A critical contribution is the inter-ticker contrastive loss that resolved representation collapse (mean cosine similarity from 0.96 to 0.24), enabling genuinely differentiated portfolio weights. Phase 3 implements lightweight personalization via a 76-parameter LoRA module that infers investment objectives from revealed trading behavior rather than questionnaires, incorporating natural language goal specification (e.g., "buy a house in 3 years"). 

\subsection{Portfolio Optimization with Explainability and Retail Accessibility}
\citet{bachhav2024ai} developed an AI-based personalized portfolio allocation engine targeting retail investors, integrating a composite risk score, market regime detection, multi-factor equity scoring, and tax-aware rebalancing with FIFO lot tracking and tax-loss harvesting. Their implementation demonstrated that tax-loss harvesting reduced simulated rebalancing tax liability by 39.8\%, with annualized returns of 8.2\%–16.1\% across risk profiles. The platform also employs Explainable AI modules to convert algorithmic decisions into plain-language narratives, addressing a key transparency gap in robo-advisory literature. 

\subsection{Foundational Concepts and Modern Adaptations}
The present work builds upon seminal contributions to portfolio theory and tax-aware investing. \citet{berkin2003tax} established foundational methods for simulating tax-loss harvesting strategies, while \citet{sialm2018tax} demonstrated that tax awareness in actively managed factor strategies primarily operates through gain deferral rather than loss maximization. The literature increasingly emphasizes the importance of accounting for regulatory constraints (wash-sale rules), market microstructure (transaction costs, bid-ask spreads), and behavioral finance considerations (disposition effects, investor heterogeneity) when implementing tax-efficient strategies. 

Recent advances integrate technological enablement with financial theory. The decline in trading costs and rise of fintech platforms have democratized tax-aware investing, previously accessible only to high-net-worth and institutional investors. Concurrently, deep learning and reinforcement learning methods enable dynamic, personalized optimization that simultaneously balances multiple objectives, alpha generation, tax efficiency, and behavioral alignment, at scale and low cost.

\section{Hypotheses}
We hypothesize that portfolio optimization performance, measured by after-tax returns, follows a clear hierarchical ordering based on architectural sophistication and information integration. We are going to conduct a 2x2 factorial experiment in which a multi-agent investment advisor system generates tax-optimized trade recommendations for simulated client portfolios under four conditions: (1) a no-augmentation baseline, (2) a custom capital-gains calculation engine only (Tax Engine), (3) a RAG-retrieved vector store of market and tax advisory reports only, and (4) both components active simultaneously. The primary outcome is the percentage of reduction in projected capital-gains tax liability relative to a no-trade baseline. Within this design, we test the following hypotheses. 

\begin{enumerate}
  \item \textbf{Baseline Underperformance}: A baseline deep reinforcement learning policy trained on pre-tax returns alone, without explicit tax reasoning or external knowledge, will deliver the poorest after-tax performance relative to all augmented conditions. 
  \item \textbf{Tax-Aware Improvement}: A tax-specialized agent trained with explicit loss-harvesting objectives, position-level tax-lot tracking, and wash-sale constraints will substantially outperform the baseline, capturing the documented tax alpha from loss harvesting while maintaining competitive pre-tax returns.
  \item \textbf{RAG Knowledge Benefit}: A retrieval-augmented policy that grounds decisions in contextual market knowledge, earnings announcements, volatility regimes, tax documentation, regulatory changes, will outperform the baseline by improving decision quality through better information access, though likely with smaller marginal gains than the tax agent alone. 
  \item \textbf{Synergistic Integration}: The combination of both tax-specialized reasoning and retrieval-augmented knowledge will outperform any single augmentation, yielding a cumulative effect in which tax reasoning provides structural optimization while RAG provides adaptive context sensitivity. We further hypothesize that this synergistic combination will approach or exceed the theoretical tax alpha established in prior literature, demonstrating that learned policies can internalize the insights from decades of tax optimization research. 

\end{enumerate}

\section{Methods}

This study used a 2×2 within-subjects factorial design to evaluate the independent and interactive effects of two system components: a tax optimization engine and a vectorized knowledge retrieval system on the quality of AI-generated tax-optimized trade recommendations. The independent variables were the tax engine (present vs. absent) and knowledge retrieval system (present vs. absent), yielding four conditions: both active, tax engine only, retrieval only, and a no-component baseline. All conditions ran on the same multi-agent AI architecture with only the relevant components toggled on or off via configuration settings, ensuring observed differences are attributable solely to those components. Each portfolio was evaluated under all four conditions, making portfolio the repeated-measures unit. A two-way repeated-measures ANOVA was selected because it estimates both main effects and their interaction within a single omnibus test, accounts for individual portfolio variability in the error term (increasing statistical power), and is more efficient than one-factor-at-a-time comparisons (\citet{montgomery2017design}).

The experimental units were 30 simulated client portfolio scenarios, sized to achieve power of $1 - \beta \geq 0.80$ for a medium effect (f = 0.25, $\alpha$ = .05; \citet{cohen1988statistical}). Each synthetic portfolio scenario represented a taxable brokerage account containing between 5 and 20 security positions, with at least one unrealized capital loss to ensure a harvestable opportunity existed. Scenarios varied in total value (ranging from \$25,000 to \$500,000) and in the ratio of unrealized gains to losses, reflecting a realistic cross-section of retail client accounts. Because this was a within-subjects design, each of the 30 portfolios was evaluated under all four conditions, yielding 120 total observations (30 portfolios × 4 conditions). Portfolio holdings, account structure, and capital gains history were held constant across conditions via deterministic seeding, ensuring that any differences in outcomes are attributable to the experimental manipulation rather than portfolio characteristics.

The main experimental loop executed each of the 30 portfolio scenarios through all four conditions. For each iteration, a LangChain-based multi-agent orchestration system was invoked. An Orchestrator Agent delegated tasks to specialized sub-agents, each with access to domain-specific tools. The agent workflow operated as follows: upon receiving a request containing account context, priority information, and user intent, the Orchestrator Agent first delegated to a Holdings Analyst to retrieve current client holdings. Next, it elected from a pool of specialized worker agents: Research, Tax, Mathematics, and Guidance agents. Each worker agent operated in parallel, with access to its respective tools: public market data APIs for pricing and fundamental data, a vector database of tax rules and regulations, a tax optimization engine for cost basis computation and tax-loss harvesting strategy generation, and a vector database of portfolio management best practices for contextual guidance. 

The factorial conditions correlate with the agents’ access to various components of the orchestration architecture, where the tax engine condition toggled access to sections of agents’ system prompts and the tax guidance condition toggled access to RAG tooling. 

The primary dependent variable was percentage tax savings; the proportional reduction in projected federal capital gains tax liability relative to a no-trade baseline:  
\[
\text{Tax Savings} (\%) = \frac{T_{\text{baseline}} - T_{\text{recommended}}}{T_{\text{baseline}}} \times 100\%
\]

\section{Results}
\subsection{Descriptive Statistics}
\begin{table}[htbp]
\centering
\caption{Descriptive Statistics for Tax Savings by Condition}
\label{tab:descriptive_stats}
\begin{tabular}{llcccc}
\toprule
\textbf{Tax Tools} & \textbf{RAG} & \textbf{n} & \textbf{M} & \textbf{SD} & \textbf{Mdn} \\
\midrule
Off & Off (Baseline) & 30 & 30.55\% & 99.16 & 24.01 \\
Off & On (RAG Only) & 30 & 47.73\% & 45.98 & 35.63 \\
On & Off (Tax Tools Only) & 30 & $-$11.03\% & 114.96 & 0.00 \\
On & On (Tax Tools + RAG) & 30 & $-$21.02\% & 161.94 & 0.00 \\
\bottomrule
\end{tabular}
\end{table}

\begin{figure}[htbp]
\centering
\includegraphics[width=0.8\textwidth]{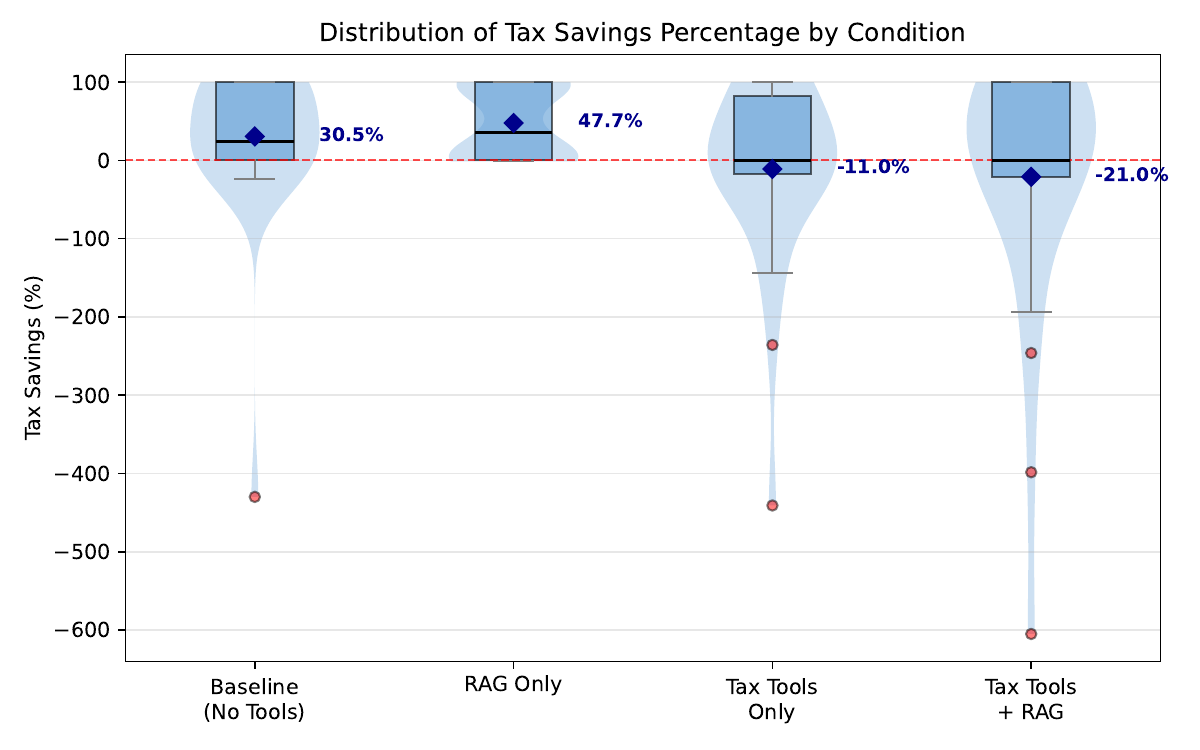}
\caption{Distribution of Tax Savings Across Experimental Conditions}
\label{fig:boxviolin}
\end{figure}

When the tax optimization engine was disabled, portfolios achieved positive mean tax savings regardless of RAG status $(30.55\%\ \text{and}\ 47.73\%)$. When the tax optimization engine was enabled, mean tax savings became negative $(-11.03\%\ \text{and}\ -21.02\%)$, indicating that the engine's trade recommendations on average increased tax liability. Variability was highest in the Tax Tools + RAG condition $(SD=161.94)$.

\subsection{Omnibus 2 × 2 Repeated-Measures ANOVA }

A two-factor repeated-measures ANOVA was conducted with tax optimization engine (on/off) and tax documents knowledge retrieval (on/off) as within-subjects factors and tax savings percentage as the dependent variable. Table 2 presents the omnibus results. 

\begin{table}[htbp]
\centering
\caption{Repeated-Measures ANOVA for Tax Savings Percentage}
\label{tab:anova_results}
\begin{tabular}{lcccccc}
\toprule
\textbf{Source} & \textbf{SS} & \textbf{df} & \textbf{MS} & \textbf{F} & \textbf{p} & ${\eta^2}_p$ \\
\midrule
Tax Tools & 91,289.80 & 1, 29 & 91,289.80 & 9.17 & .005 & .240 \\
RAG & 389.09 & 1, 29 & 389.09 & 0.04 & .841 & .001 \\
Tax Tools $\times$ RAG & 5,536.84 & 1, 29 & 5,536.84 & 0.36 & .553 & .012 \\
\bottomrule
\end{tabular}
\end{table}

\begin{figure}[htbp]
\centering
\includegraphics[width=0.8\textwidth]{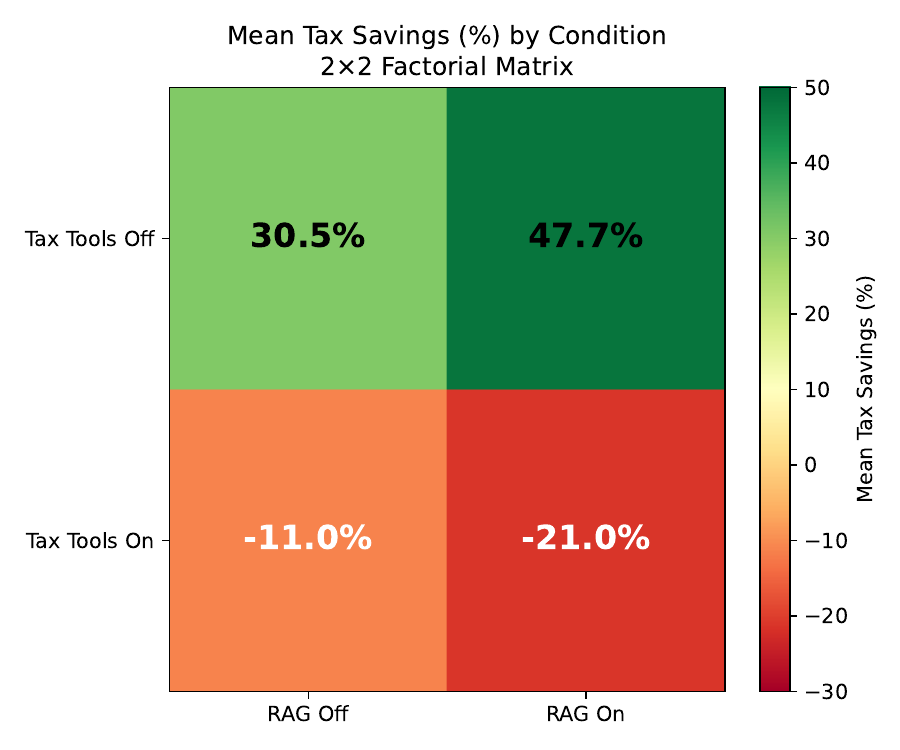}
\caption{Distribution of Tax Savings Across Experimental Conditions}
\label{fig:boxviolin}
\end{figure}

Sphericity was not violated ($\epsilon = 1.0$ for all effects in this 2-level design; Greenhouse-Geisser correction is unnecessary with $df = 1$).
The main effect of the tax optimization engine was significant, $F(1, 29) = 9.17, p = .005, \eta^2_p$ = .240, indicating a large effect. Portfolios evaluated with the tax engine enabled $(M = -16.03\%)$ achieved significantly lower tax savings than those without it $(M = 39.14\%)$. The main effect of RAG was not significant, $F(1, 29) = 0.04, p = .841, \eta^2_p = .001$. The two-way interaction was not significant, $F(1, 29) = 0.36, p = .553, \eta^2_p$ = .012, indicating that the detrimental effect of the tax engine did not depend on whether RAG was also enabled. Because the interaction was not significant, main effects are interpreted directly. 

\subsection{Post-Hoc Pairwise Comparisons}
Bonferroni-corrected pairwise comparisons were conducted to decompose the significant main effect of the tax optimization engine. Table 3 presents the within-subjects pairwise results. 

\begin{table}[htbp]
\centering
\caption{Bonferroni-Corrected Pairwise Comparisons }
\label{tab:contrasts}
\begin{tabular}{llcccc}
\toprule
\textbf{Contrast} & \textbf{Comparison} & \textbf{t(29)} & \textbf{p corrected} & \textbf{Hedges' g} \\
\midrule
Tax Tools (main effect) & Off vs. On & 3.03 & .005 & 0.67 \\
RAG (main effect) & Off vs. On & $-$0.20 & .841 & $-$0.04 \\
Interaction: Tax Tools = Off & RAG Off vs. On & $-$1.04 & .618 & $-$0.22 \\
Interaction: Tax Tools = On & RAG Off vs. On & 0.27 & 1.000 & 0.07 \\
\bottomrule
\end{tabular}
\end{table}

The significant main effect of Tax Tools $(t(29) = 3.03,\ p_{bonf} = .005,\ g = 0.67)$ reflects a medium-to-large effect: disabling the tax optimization engine yielded approximately 55 percentage points more tax savings than enabling it. No other comparisons reached significance after Bonferroni correction.

\subsection{Sensitivy Analysis}

To assess robustness to extreme values, outliers on tax savings percentage were identified using the IQR method (bounds: [-150.00\%, 250.00\%]). Because the design is within-subjects, entire portfolios were excluded if any of their four observations fell outside these bounds. Seven portfolios were excluded, leaving 23 portfolios (92 observations).

\begin{table}[htbp]
\centering
\caption{Sensitivity Analysis RM-ANOVA After Outlier Exclusion (N = 23) }
\label{tab:sensitivity_anova}
\begin{tabular}{lccc}
\toprule
\textbf{Source} & \textbf{F(1, 22)} & \textbf{p} & $\eta^2_p$ \\
\midrule
Tax Tools & 4.38 & .048 & .166 \\
RAG & 1.62 & .216 & .069 \\
Tax Tools $\times$ RAG & 0.23 & .635 & .010 \\
\bottomrule
\end{tabular}
\end{table}

The pattern of results was unchanged: the tax optimization engine main effect remained significant (p = .048), while RAG and the interaction remained non-significant. The effect size was attenuated ($\eta^2_p = .166 vs. .240$) due to the removal of extreme cases, but the direction and conclusion are consistent with the primary analysis. 

\subsection{Supplementary Analysis: Tax Savings Amount }
A parallel 2 × 2 repeated-measures ANOVA was conducted on tax savings amount (dollars) as a secondary dependent variable.

\begin{table}[htbp]
\centering
\caption{Repeated-Measures ANOVA for Tax Savings Amount (\$)}
\label{tab:robustness_anova}
\begin{tabular}{lccc}
\toprule
\textbf{Source} & \textbf{F(1, 29)} & \textbf{p} & $\eta^2_p$ \\
\midrule
Tax Tools & 5.47 & .026 & .159 \\
RAG & 0.02 & .890 & .001 \\
Tax Tools $\times$ RAG & 1.14 & .294 & .038 \\
\bottomrule
\end{tabular}
\end{table}

The dollar-value analysis converged with the percentage-based findings: the tax optimization engine significantly reduced tax savings in absolute terms (p = .026), while RAG and the interaction were not significant. 

\subsection{Non-Parametric Robustness Check }

Given the high variability and non-normal distribution of tax savings percentage, a Friedman test was conducted as a non-parametric alternative to the omnibus ANOVA. The Friedman test was significant: $\chi^2(3) = 10.47, p = .015, W = 0.116$, confirming that the conditions differed significantly in tax savings when no distributional assumptions are imposed. This is consistent with the parametric findings. 

\subsection{Summary}
Across all analyses the primary 2 × 2 repeated-measures ANOVA, the sensitivity analysis with outlier exclusion, the supplementary dollar-value analysis, and the non-parametric Friedman test, results converge on a single conclusion: enabling the tax optimization engine significantly reduced tax savings (or increased tax liability), while RAG knowledge retrieval had no measurable effect and did not interact with the tax engine. The effect of Tax Tools was large $(\eta^2_p = .240, g = 0.67)$, robust to outlier exclusion, and consistent across both percentage and dollar-denominated outcomes.  

\section{Conclusion} 
The results largely contradicted our initial hypotheses. The omnibus ANOVA revealed a significant main effect of the tax optimization engine, $F(1, 29) = 9.17, p = .005, \eta^2_p = .240$, representing a large effect. Portfolios evaluated with the tax engine enabled produced mean tax savings of -16.0\%, compared to +39.1\% when the engine was disabled a difference of approximately 55 percentage points. The main effect of RAG was not significant, F(1, 29) = 0.04, p = .841, and the Tax Tools × RAG interaction was not significant, F(1, 29) = 0.36, p = .553, indicating that the detrimental effect of the tax engine did not depend on whether RAG was also enabled. This pattern was confirmed by a non-parametric Friedman test, $\chi^2(3) = 10.47, p = .015$, and held under a sensitivity analysis excluding outlier portfolios (p = .048). 

Rather than underperforming all augmented conditions, the Baseline agent achieved a mean tax savings of 30.5\%, suggesting that the pre-trained language model's internalized financial knowledge is sufficient to generate competent tax-loss harvesting recommendations without explicit tooling. The RAG Only condition performed best (47.7\%), though this advantage over Baseline was not statistically significant after Bonferroni correction. Most critically, both conditions with the tax engine enabled produced negative mean tax savings, Tax Tools Only (-11.0\%) and Tax Tools + RAG (-21.0\%) meaning the engine's recommendations on average increased tax liability. Bonferroni-corrected post-hoc comparisons confirmed the significant Tax Tools main effect $(t(29) = 3.03, p_{bonf} = .005, Hedges' g = 0.67)$, a medium-to-large effect. 

These findings carry several important limitations. All scenarios were synthetically generated and evaluated under a single market regime, limiting ecological validity. The study did not verify compliance with wash-sale rules or assess the qualitative soundness of individual trade recommendations beyond aggregate tax impact. Additionally, while the within-subjects design increased statistical power by controlling for portfolio-level variability, the high within-condition variance (SDs ranging from 46\% to 162\%) suggests substantial heterogeneity in how the agent handles different portfolio configurations. 

Despite these limitations, the results offer practical insight for developers of AI-driven wealth management systems: augmenting language model agents with domain-specific tooling does not guarantee improved performance and may in fact degrade it if the integration introduces friction or conflicting optimization signals. The tax engine's negative impact likely reflects over-constrained decision-making, where rule-based cost-basis calculations conflicted with the agent's broader reasoning about optimal trade selection. Future work should examine larger and more diverse portfolio samples, incorporate real client data under multiple market regimes, refine the tax engine's integration to complement rather than constrain agent reasoning, and evaluate recommendation quality along dimensions beyond tax savings alone including wash-sale compliance, portfolio risk characteristics, and alignment with individual investor objectives. 

\bibliography{references}
\end{document}